\documentclass{article}

\usepackage{kafkanotes}
\usepackage{bm}
\usepackage{cleveref}
\usepackage{natbib}
\usepackage{amssymb}
\usepackage{amsmath}
\DeclareMathOperator*{\argmax}{arg\,max}

\usepackage{cancel}
\usepackage{xcolor}
\newcommand\Ccancel[2][black]{\renewcommand\CancelColor{\color{#1}}\cancel{#2}}

\title{Denoising Diffusion Probabilistic Models in Six Simple Steps}
\author{Richard E.~Turner\textsuperscript{a,b}, Cristiana-Diana Diaconu\textsuperscript{a}, Stratis Markou\textsuperscript{a}, Aliaksandra Shysheya\textsuperscript{a},\\ Andrew Y.~K.~Foong\textsuperscript{b} and Bruno Mlodozeniec\textsuperscript{a}\\ \textsuperscript{a} Department of Engineering, University of Cambridge, UK\\ \textsuperscript{b} Microsoft Research, Cambridge, UK\\}

\begin{document}

\begin{titlepage}
\thispagestyle{empty}
\maketitle

%Abstract
\begin{abstract}
Denoising Diffusion Probabilistic Models (DDPMs) \citep{ho+al:2020} are a very popular class of deep generative model that have been successfully applied to a diverse range of problems including image and video generation, protein and material synthesis, weather forecasting, and neural surrogates of partial differential equations. 
Despite their ubiquity it is hard to find an introduction to DDPMs which is simple, comprehensive, clean and clear. The compact explanations necessary in research papers are not able to elucidate all of the different design steps taken to formulate the DDPM and the rationale of the steps that are presented is often omitted to save space. Moreover, the expositions are typically presented from the variational lower bound perspective which is unnecessary and arguably harmful as it obfuscates why the method is working and suggests generalisations that do not perform well in practice.  On the other hand, perspectives that take the continuous time-limit are beautiful and general, but they have a high barrier-to-entry as they require background knowledge of stochastic differential equations and probability flow. 
In this note, we distill down the formulation of the DDPM into six simple steps each of which comes with a clear rationale. We assume that the reader is familiar with fundamental topics in machine learning including basic probabilistic modelling, Gaussian distributions, maximum likelihood estimation, and deep learning.

\end{abstract}

%\tableofcontents
\end{titlepage}

%Geometri untuk halaman konten
\newgeometry{top=20mm,bottom=25mm,right=80mm,left=20mm}

%================KONTEN DIMULAI DISINI================%

\section{Preliminaries}

Let's start by defining the problem that we're interested in solving.

\textbf{Problem definition}. We have a large amount of training data $x$ that come from an underlying distribution $q(x)$.\sidenote{\footnotesize To simplify the notation we will consider one dimensional real-valued data, but the generalisation to $D$ dimensional data is straightforward. We will also assume that the data are zero mean and unit variance.} We want to fit a probabilistic model $p(x)$ to this training data\sidenote{\footnotesize In practice a model with latent variables $z$  will be used $p(x,z)$ so $p(x)$ is specified implicitly through an intractable marginalisation $p(x) = \int p(x,z)\;\mathrm{d}z $.} which will approximate $q(x)$ with the main aim being to sample new data $x \sim p(\cdot)$.\sidenote{\footnotesize Note that we will want $p(x)$ to be a better approximation of $q(x)$ than the raw data distribution $\frac{1}{N}\sum_{n=1}^N \delta(x-x_n)$ -- so samples from $p(x)$ should not simply regurgitate the training data and should be plausible new samples from $q(x)$.}  

\section{The Six Steps of the DDPM}

We now break down the DDPM approach into six simple steps, each with a clear rationale and an associated design-space.

\subsection{Augmentation}

The goal of the first step of the DDPM is to turn the hard unsupervised generative modelling problem into a series of simple supervised regression problems. We can then leverage standard deep learning tools for supervised regression, which interpolate and generalise well, in order to learn a generative model. 

\begin{figure}[!h]\centering 
 \includegraphics[scale = 0.35]{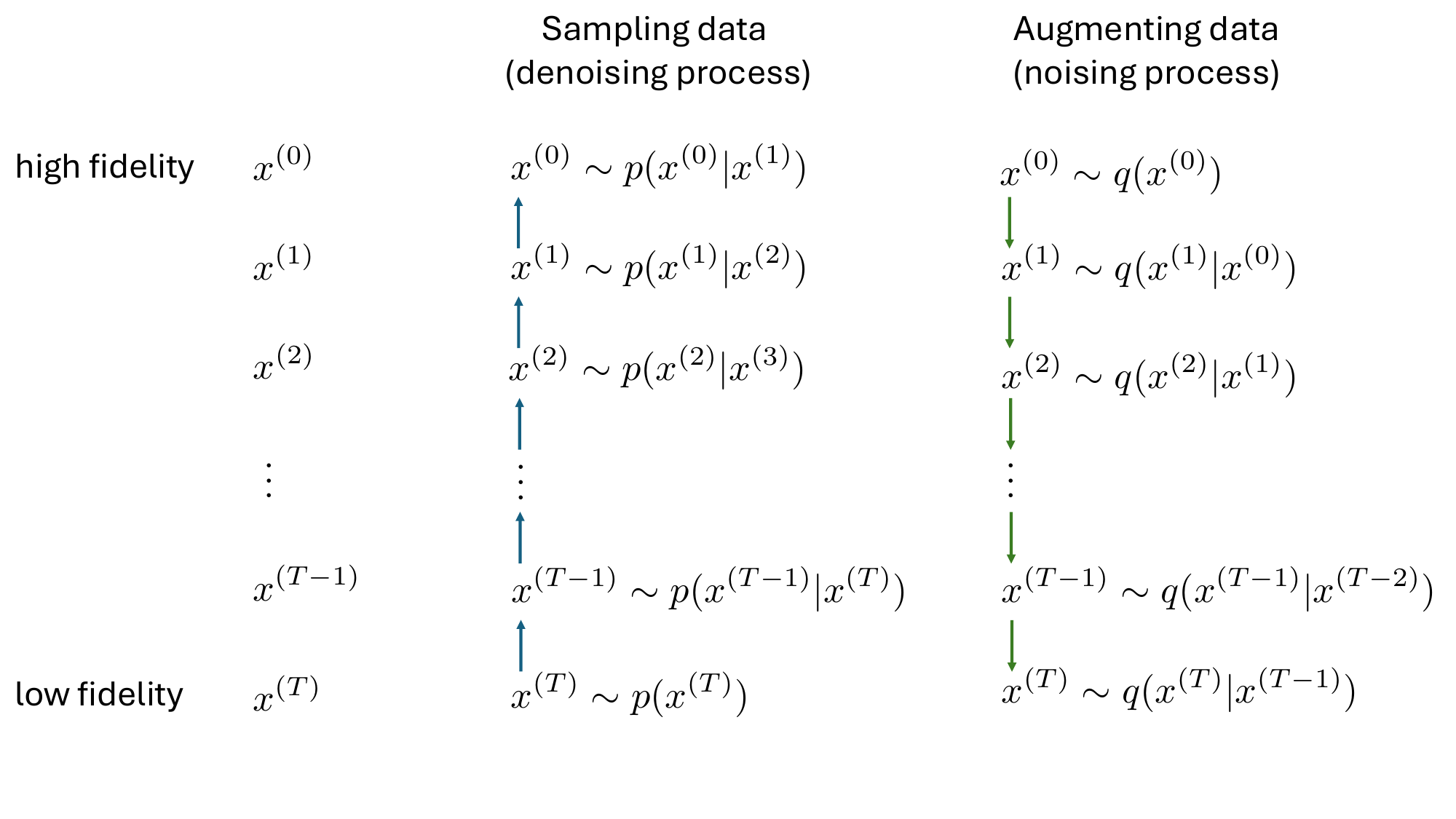}
 \caption{The augmented data at different fidelity levels (left), the data generation or denoising process used at test time (centre), and the augmentation process that generates the different fidelity levels via the noising process for training (right).}
 \label{fig:noising-and-denoising}
 \end{figure}

\textbf{The Augmentation Conditions}. The conversion from a generative modelling problem into a supervised learning problem is achieved by augmenting the original training data with $T$ additional fidelity levels. Specifically we'll denote the augmented data as 
\[ 
x^{(0:T)} = [ x^{(0)}, x^{(1)},x^{(2)}, \hdots, x^{(T-1)},  x^{(T)} ]
\]
and we will set up the augmentation to meet the following conditions
\begin{enumerate}
\item the highest fidelity data $x^{(0)}$ will be the original training data (by construction),
\item the lowest fidelity level will be simple to sample from directly i.e.~$p(x^{(T)})$ will have a simple form,
\item predicting a higher fidelity level from the fidelity level below it will be a simple regression problem i.e.~$p(x^{(t-1)} | x^{(t)})$ is easy to model and learn.\sidenote{\footnotesize One worry is that although each regression problem is simple to model and learn, small errors made at each stage could compound into large errors by the time we generate the data. However, in certain cases theory is available which guarantees that this does not happen \citep{benton+al:2024,chen+al:2023a,chen+al:2023b}. These papers bound the KL divergence between the data distribution and model marginal for certain augmentation processes, showing that the KL can be made arbitrarily small if a sufficient number of augmentations $T$ are used.}
\end{enumerate}
If the augmentation obeys these conditions, then we can generate data by sampling from the lowest fidelity $x^{(T)} \sim p(x^{(T)})$ and then recursively sample up the fidelity levels $x^{(t-1)} \sim p(x^{(t-1)} | x^{(t)})$ for $t=T \hdots 1$ until we generate $x^{(0)}$ (see the centre panel of fig.~\ref{fig:noising-and-denoising}).\sidenote{\footnotesize In this construction the augmentations $x^{(1:T)}$ can be considered to be latent variables so that \[ p(x^{(0)}) = \int p(x^{(0:T)}) \mathrm{d} x^{(1:T)}. \]}\sidenote{\footnotesize \textbf{Relationship to neural auto-regressive models}. Neural auto-regressive models \citep{larochelle+murray:2016} construct a generative model by using the product rule to decompose a joint distribution into a set of low-dimensional conditional distributions 
\[p(\mathbf{x}) = p(x_1)p(x_2 |x_1)p(x_3 | x_{1:2}) \hdots p(x_D | x_{1:D-1}).\]
These low-dimensional conditional distributions $p(x_d|x_{1:d-1})$ are now simple regression problems with inputs $x_{1:d-1}$ and targets $x_d$ for which neural networks are well suited. In this context, we can consider our setup as defining a neural auto-regressive model on an augmented dataset \begin{align}p(x^{(0:T)})&=p(x^{(T)})\prod_{t=1}^Tp(x^{(t-1)} | x^{(t:T)}) \nonumber\\
&=p(x^{(T)})\prod_{t=1}^Tp(x^{(t-1)} | x^{(t)})
. \nonumber\end{align} Here we have used the fact that as the noising process is first order Markov, the optimal denoising process is also first order Markov and so longer range dependencies can be discarded. 
In the multi-dimensional case, blocks of variables $x^{(t-1)}$ (rather than single scalar variables) are generated in each auto-regressive step.}

\textbf{Meeting the Augmentation Conditions}. A simple augmentation strategy, which we will show satisfies the conditions above, is to sample a training data point at random, $x^{(0)} \sim q(x^{(0)})$ (thereby satisfying property 1) and to progressively add more and more noise to the data point to produce progressively coarser fidelity levels (see the right panel of fig.~\ref{fig:noising-and-denoising}). 

More generally, we can construct a lower fidelity level from the one above by multiplying by a coefficient $\lambda_t$ (that will typically be close to 1) and adding Gaussian noise of variance $\sigma_t^2$ so\sidenote{\footnotesize I am using standard notation for Gaussian AR processes, but this differs from the standard notation in diffusion which has $x^{(t)} = \sqrt{\alpha_t} x^{(t-1)} + \sqrt{\beta}_t \epsilon_t$. I.e.~
$\alpha_t = \lambda_t^2$ and 
$\beta_t$ = $\sigma_t^2$. For a full mapping between the notation used here and that in e.g.~\cite{ho+al:2020} see appendix \ref{sec:notation}.}
\[ x^{(t)} = \lambda_t x^{(t-1)} + \sigma_t \epsilon_t \;\; \text{where} \;\epsilon_t \sim \mathcal{N}(0,1).
 \]
This noising process is called a first-order Gaussian auto-regressive process\sidenote{\footnotesize \emph{First order} because each fidelity level only depends on the previous one (a first order Markov structure); \emph{Gaussian} because of the noise distribution; and \emph{auto-regressive} as it looks like a linear regression model where the inputs were previous outputs in the sequence.} and it can equivalently be written in distribution form
\[ q(x^{(t)} | x^{(t-1)}) = \mathcal{N}(x^{(t)} ; \lambda_{t} x^{(t-1)},\sigma_t^2). \]
The parameters of the augmentation process, $\{\lambda_t,\sigma_t^2 \}_{t=1}^T$ and the number of augmentation steps $T$, need to be selected to fulfill conditions 2 and 3 above. Luckily the use of a linear-Gaussian augmentation strategy means that we can use analytic results to figure out how to do this.

It turns out that there is some freedom, but one popular choice first sets $\sigma_t^2 = 1 - \lambda_t^2$ which ensures that marginally the augmentations have zero mean and unit variance i.e.~$\mathbb{E}_{q(x^{(t)})}[x^{(t)}] = 0$ and $\mathbb{E}_{q(x^{(t)})}[(x^{(t)})^2] = 1$.\sidenote{\footnotesize To prove this, consider how the mean and variance of the previous state are affected when passing through the linear Gaussian dynamics. A full derivation can be found in appendix \ref{sec:variance-preserve}.} This is known as a variance preserving augmentation and in this case the conditional distribution of an augmentation at level $t$ given the data $x^{(0)}$ is given by\sidenote{\footnotesize This can be shown by unrolling the auto-regressive dynamics. A full derivation can be found in appendix \ref{sec:condition-init}.} 
\begin{align}
q(x^{(t)}|x^{(0)}) = \mathcal{N}\left(x^{(t)}; \left( \prod_{t'=1}^t \lambda_{t'} \right) x^{(0)}, 1 - \prod_{t'=1}^t \lambda_{t'}^2 \right)\label{eq:cond-samp}.
\end{align}
Notice that if we have lots of levels ($T \rightarrow \infty$) and select $\lambda_t < 1$ then we will forget the initial condition (since $\prod_{t'=0}^T \lambda_{t'} \rightarrow 0$) and then the marginal distribution of the lowest fidelity state takes a simple form  $q(x^{(T)}) \rightarrow \mathcal{N}(x^{(T)}; 0,1) = p(x^{(T)})$ which is a very easy distribution to sample from. So condition 2 can be approximately satisfied by having a large number of levels.

Finally, we can satisfy condition 3 by noticing that if we add only a small amount of noise between levels (equivalently $\lambda_t \lesssim 1$ so that $\sigma_t^2 = 1 - \lambda_t^2 \gtrsim 0 $) then performing the regression problem $p(x^{(t-1)} | x^{(t)})$  -- which intuitively involves removing this small amount of noise from  $x^{(t)}$ to estimate $x^{(t-1)}$ and providing an uncertainty estimate --  will be simple (indeed in the limit $\lambda_t \rightarrow 1$ the mapping is the identity $p(x^{(t-1)} | x^{(t)}) \rightarrow \delta(x^{(t)} - x^{(t-1)})$ ).\sidenote{\footnotesize Note that condition 2 (the limiting distribution should be simple to sample from) and condition 3 (the denoising process should be simple to model and learn) are in tension with one another. On the one hand, for a fixed number of fidelity levels $T$, using $\lambda_t \approx 0$ leads to limiting distributions that are very close to a standard Gaussian, but they make denoising much harder as lots of noise is added in each step. On the other hand using $\lambda_t \approx 1$ makes denoising simpler, but the limiting distribution is further from a standard Gaussian.} 

\subsection{Supervised Step-wise Objective Function}

We will initially consider the case where each of the regression problems has its own individual set of parameters $\theta_{t-1}$ i.e.~$p(x^{(t-1)} | x^{(t)},\theta_{t-1})$. A simple way to train each regression model is to treat the augmented dataset just like a regular dataset and perform maximum-likelihood learning of the parameters 
\[
\theta_{0:T-1}^*= \argmax_{ \theta_{0:T-1}} \; \mathcal{L}(\theta_{0:T-1}) 
\]
where the log-likelihood of the parameters is \[ \mathcal{L}( \theta_{0:T-1})=\mathbb{E}_{q(x^{(0:T)})}[ \log p(x^{(0:T)}|\theta_{0:T-1})]. \]
I.e.~we find the parameters $\theta_{0:T-1}$ that make the augmented dataset as probable as possible.\sidenote{\footnotesize Practically the expectation over $q(x^{(0:T)})$ will be performed by averaging over lots of samples from our augmented training data set $x_n^{(0:T)} \sim q(x^{(0:T)})$: \begin{align} \mathcal{L}(\theta_{0:T-1}) &= \mathbb{E}_{q(x^{(0:T)})}[ \log p(x^{(0:T)}|\theta_{0:T-1})] \nonumber \\
&  \approx \frac{1}{N} \sum_{n=1}^N  \log p(x_n^{(0:T)}|\theta_{0:T-1}) \nonumber \end{align}}

We can now substitute in the form of the model to simplify the maximum likelihood objective
\begin{align}
\mathcal{L}(\theta_{0:T-1}^* ) %&= \mathbb{E}_{q(x^{(0:T)})}[ \log p(x^{(0:T)}|\theta_{1:T})] \nonumber\\
& =\mathbb{E}_{q(x^{(0:T)})}\left[ \log \left (  p(x^{(T)}) \prod_{t=1}^T p(x^{(t-1)}|x^{(t)},\theta_{t-1}) \right)\right] \nonumber\\
& =\mathbb{E}_{q(x^{(T)})}[ \log  p(x^{(T)})] + \sum_{t=1}^T \mathbb{E}_{q(x^{(t-1)},\;x^{(t)})} [ \log p(x^{(t-1)}|x^{(t)},\theta_{t-1})]. \nonumber
\end{align}
Note that only the second term on the right hand side depends on the parameters and that each $\theta_t$ appears in a single term in the sum. So
\begin{align}
\theta_{t-1}^* &= \argmax_{\theta_{t-1}} \; \mathcal{L}_{t-1}(\theta_{t-1}) \;\;\text{where} \nonumber \\
\mathcal{L}_{t-1}(\theta_{t-1}) &= 
\mathbb{E}_{q(x^{(t-1)},\;x^{(t)})}[ \log p(x^{(t-1)} | x^{(t)},\theta_{t-1})]. \nonumber
\end{align}
Which involves simply fitting a regression model by maximum likelihood learning that maps between each fidelity level in the augmented training data set.\sidenote{\footnotesize \textbf{Relationship to variational auto-encoders}. We have previously mentioned that our model can be considered to be a latent variable model with latents $x^{(1:T)}$. The Evidence Lower Bound (ELBO) is often used to perform learning and inference in such models \citep{kingma+welling:2022}. It turns out that the DDPM can be presented in this way when the approximate posterior is fixed (not learned) and corresponds to the noising distribution. In this way it can be seen as an instance of a variational auto-encoder. This perspective, and the reasons why we do not think it should have a central place in the exposition, is explained in appendix \ref{sec:ELBO}.}

\subsection{Parameter Tying and a Joint Objective}

Typically diffusion models have large numbers of steps $T$. If we want to use flexible neural networks to parameterise each conditional distribution with $K$ parameters each, then this leads to an explosion in the number of parameters (there will be $T \times K$ of them) and a large memory cost from instantiating the different models. In order to retain flexibility with a small set of parameters, we will instead share parameters across each of the regression problems. A simple way to do this is to build models that amortize across fidelity levels i.e.~the model takes in the fidelity level $t-1$, a shared set of parameters $\theta$, and the previous level's variables $x^{(t)}$ and outputs a distribution over the next fidelity level's variables  $p(x^{(t-1)} | x^{(t)},\theta,t-1)$.\sidenote{\footnotesize One example of amortization is in image modelling where $x^{(0)}$ is an image. Here a convolutional neural network called a UNet is often used to map between levels, 
$x^{(t-1)} = \text{UNet}(x^{(t)};\theta_{t-1})$.
The level-specific parameters of the UNet $\theta_{t-1}$, the convolutional filters, are produced by modulating a global set of parameters $\theta$ in a level-dependent way $\theta_{t-1} = \text{FiLM}(\theta; $t-1$))$. The $\text{FiLM}$  modulation simply scales and shifts each parameter so $\theta_{i,t-1} = \kappa_{i}(t-1) \theta_{i} + \delta_{i}(t-1)$.  The scale $\kappa_{i}(t-1)$ and the shift $\delta{i}(t-1)$ are themselves produced using a multi-layer perceptron which perfoms the amortisation by taking $t-1$ as input in the form of a sinusoidal encoding.}

\begin{marginfigure}%
\includegraphics[width=1\linewidth]{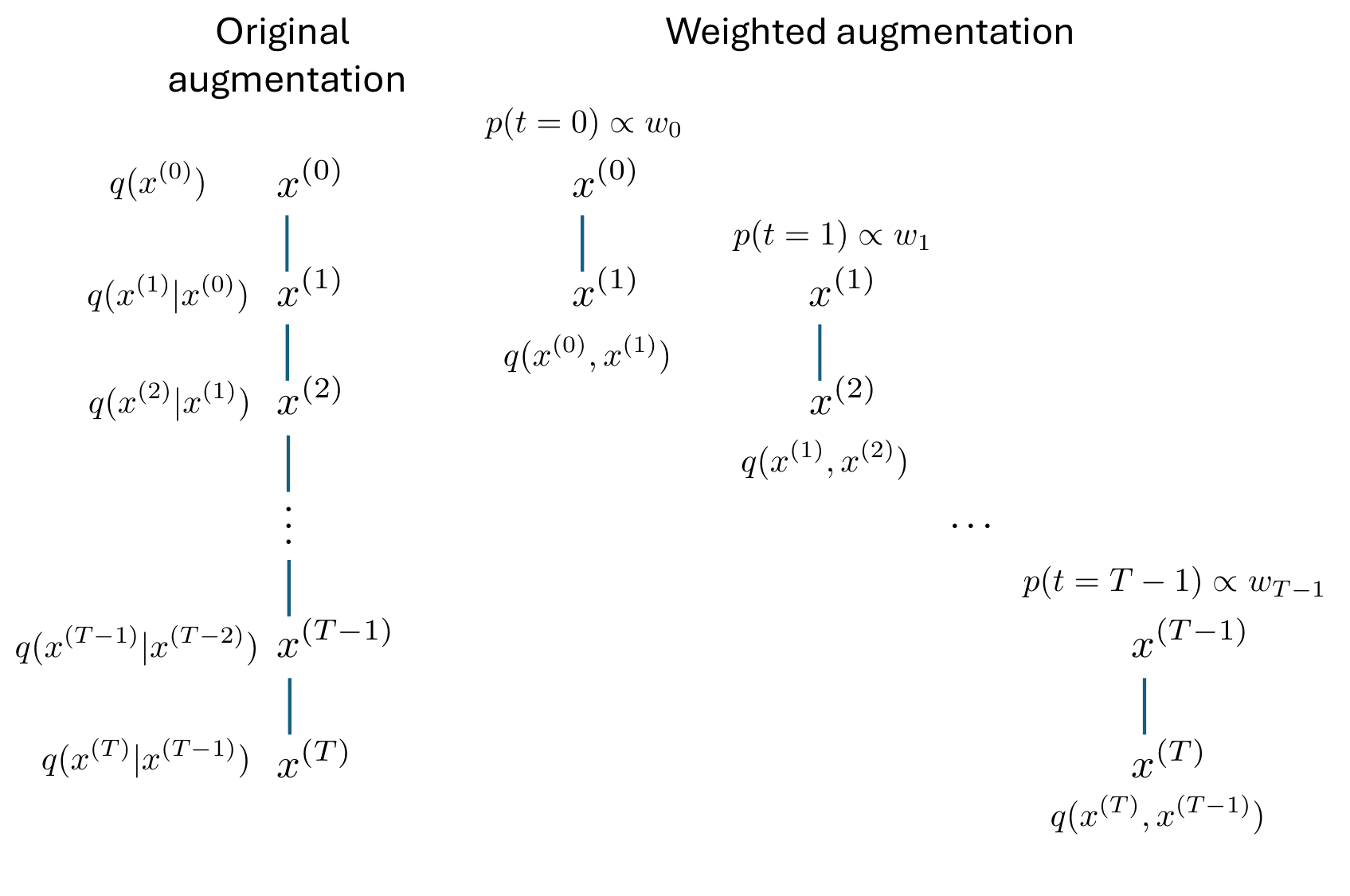}
   \caption{Weighted augmentation. The original augmentation scheme (left) starts by sampling a data point at random and then generates $T$ progressively noisier versions. In the weighted augmentation scheme (right), we sample a fidelity level in proportion to its weight $w_t$ and then sample pairs of points from adjacent fidelity models from the marginal. In this way, some levels will have more data points than others and will make a larger contribution to the training loss.}
   \label{fig:weighted-augmentation}
 \end{marginfigure}

As the parameters are shared across levels, the maximum-likelihood training objective no longer decomposes. Typically weights $\{ w_t \}_{t=0}^{T-1}$ are introduced to control the contribution of each fidelity level to the likelihood,\sidenote{\footnotesize As we will see later, often a Gaussian distribution is used for $p(x^{(t-1)} | x^{(t)},\theta,t-1)$ and only the mean is learned. In this case the weights allow the user to have independent control over the width of the conditional distributions during generation and the way each level trades-off against one another in the cost. If the variances are learned, it is not clear whether it is necessary to use these weights i.e.~the maximum likelihood setting $w_t=1$ might be sufficient and even when the variances are fixed some authors use the unweighted objective \citep{kingma+al:2021}}  with the pure joint maximum-likelihood training objective recovered when $w_t=1$.\sidenote{\footnotesize \label{sidenote:reweighting}Alternatively, these weights can be interpreted as performing unweighted maximum-likelihood learning on a modified augmented data set. See figure \ref{fig:weighted-augmentation} for a description of this modified augmentation procedure.}
\begin{align}
\theta^* & = \argmax_{\theta} \sum_{t=1}^T w_{t-1}\; \mathcal{L}_{t-1}(\theta)\;\; \text{where}\nonumber\\
\mathcal{L}_{t-1}(\theta) & =\mathbb{E}_{q(x^{(t-1)},\;x^{(t)})}[ \log p(x^{(t-1)} | x^{(t)},\theta,t-1)]. \nonumber
\end{align}
This new training objective appears expensive to compute at first sight as each gradient computation requires computation of $T$ terms each of which involves a forward pass through a neural network. However, stochastic optimisation saves the day. We rewrite the sum over the levels $t$ as an expectation over a uniform distribution over the integers $1 \hdots T$
\[
\frac{1}{T}\sum_{t=1}^T w_{t-1}\; \mathcal{L}_{t-1}(\theta) = \mathbb{E}_{t\sim\text{Uniform(1,T)}} \left [ w_{t-1}\;\mathcal{L}_{t-1}(\theta) \right].
\]
Now, at each step, we can sample a level $t$ at random and compute the gradient just on this level to form an unbiased estimate of the full gradient, in much the same way as we subsample training data in stochastic gradient ascent. This computation is as expensive as a single step of gradient-based learning for a single level in the original untied model. Even better, practically, this new model can learn faster than the untied model as the parameter tying means that information from level $t$ helps learn better models for $t' \ne t$. For example, the regression problems at adjacent levels $t+1$ and $t-1$ will be very similar. In this way, diffusion becomes very scalable and quick to train. 

\subsection{Gaussian Regression Model: Variance Reduction}
We now need to select a model to perform the regression. Since the noising process is Gaussian and only a very small amount of noise is added at each step, a common choice for the denoising process is to also use a Gaussian\sidenote{\footnotesize Amazingly, the optimal regression in a sensibly defined limit as $T \rightarrow \infty$ has this same form \citep{anderson:1983,song+al:2021}, justifying the choice of this family.}
\[p(x^{(t-1)} | x^{(t)},t-1,\theta) = \mathcal{N}\left(x^{(t-1)} ; \mu_{\theta}(x^{(t)},t-1),\sigma_{\theta}^2(x^{(t)},t-1) \right).\]
Here the mean of the Gaussian $\mu_{\theta}(x^{(t)},t-1)$ and the variance $\sigma_{\theta}^2(x^{(t)},t-1)$ depend on the data at the previous fidelity level $x^{(t)}$ in a non-linear way and will be specified using a neural network.

Choosing this simple regression model has another crucial advantage. It allows us to reduce the Monte Carlo noise in the training objective by performing one of the averages over the augmented training data analytically rather than by sampling.\sidenote{\footnotesize This trick relates to Rao-Blackwellisation and the local reparameterisation trick for Bayesian neural networks} This trick means we effectively train the model on an infinite set of augmentations.

To see how to do this, we first compute the key term in the maximum-likelihood objective function using this model class
%
% \begin{align} 
% &\mathbb{E}_{q(x^{(t-1)},\;x^{(t)})}[ \log p(x^{(t-1)} | x^{(t)},\theta,t-1)] = \nonumber \\
% & \quad \quad \quad - \frac{1}{2} \mathbb{E}_{q(x^{(t-1)},\;x^{(t)})} \Biggl[ \frac{1}{\sigma_{\theta}^2(x^{(t)},t-1)}(x^{(t-1)} - \mu_{\theta}(x^{(t)},t-1))^2 \nonumber \\
% & \quad \quad \quad \quad \quad \quad\quad \quad \quad \quad \quad \quad\quad \quad \quad \quad \quad \quad\quad \quad \quad   + \log 2 \pi  \sigma_{\theta}^2(x^{(t)},t-1) \Biggl]\nonumber
% \end{align}
%
\begin{align} 
&\mathcal{L}_{t-1}(\theta) =  - \frac{1}{2} \mathbb{E}_{q(x^{(t-1)},\;x^{(t)})} \Biggl[\underbrace{ \frac{(x^{(t-1)} - \mu_{\theta}(x^{(t)},t-1))^2 }{\sigma_{\theta}^2(x^{(t)},t-1)}  + \log 2 \pi  \sigma_{\theta}^2(x^{(t)},t-1) }_{f(x^{(t-1)},\;x^{(t)})}\Biggl]\nonumber
\end{align}
Practically, the expectations with respect to $q(x^{(t-1)},\;x^{(t)})$ can be computed using samples from our augmented data set. A na\"{i}ve approach samples a data point from $x^{(0)} \sim q(x^{(0)})$ and then samples the augmented versions from $x^{(t-1)},\;x^{(t)} \sim q(x^{(t-1)},\;x^{(t)} | x^{(0)})$ (which is a simple Gaussian distribution). However, a smarter approach performs the expectation over $x^{(t-1)}$ analytically using the law of nested conditional expectations
\begin{align}
\mathbb{E}_{q(x^{(t-1)},\;x^{(t)})}(f(x^{(t-1)},\;x^{(t)})) & = \mathbb{E}_{q(x^{(0)}, \;x^{(t-1)},\;x^{(t)})}[ f(x^{(t-1)},\;x^{(t)})] \nonumber\\
& = \mathbb{E}_{q(x^{(0)},\;x^{(t)})} [ \mathbb{E}_{q(x^{(t-1)}|x^{(0)},\;x^{(t)})}[f(x^{(t-1)},\;x^{(t)})]]\nonumber
\end{align}
The inner expectation here is analytic when the noising and denoising processes are Gaussian yielding\sidenote{\footnotesize The formula for the inner expectation is closely related to the KL divergence between two Gaussians.}
\begin{align} 
&\mathcal{L}_{t-1}(\theta) =  - \frac{1}{2} \mathbb{E}_{q(x^{(0)},\;x^{(t)})} \Biggl[ \frac{(\mu_{t-1|0,t} - \mu_{\theta}(x^{(t)},t-1))^2 + \sigma^2_{t-1|0,t} }{\sigma_{\theta}^2(x^{(t)},t-1)}  \nonumber \\ 
&\quad \quad \quad \quad \quad \quad \quad\quad \quad \quad \quad \quad \quad \quad\quad \quad \quad \quad \quad \quad \quad + \log 2 \pi  \sigma_{\theta}^2(x^{(t)},t-1) \Biggl]\nonumber
\end{align}
Here $q(x^{(t-1)}|x^{(0)},\;x^{(t)}) = \mathcal{N}(x^{(t-1)}; \mu_{t-1|0,t}, \sigma^2_{t-1|0,t})$ where the mean is a linear combination of $x^{(0)}$ and $x^{(t)}$, that is  $\mu_{t-1|0,t} = a^{(t-1)}  x^{(0)} + b^{(t-1)} x^{(t)}$.\sidenote{\footnotesize The full expressions for $\mu_{t-1|0,t}$ and $\sigma^2_{t-1|0,t}$ are derived in appendix \ref{sec:condition-init+fid} and shown to be
\begin{align}
   \mu_{t-1|0,t} &= a^{(t-1)}  x^{(0)} + b^{(t-1)} x^{(t)} \nonumber\\ 
   a^{(t-1)}  & = \frac{\left(\prod_{t'=1}^{t-1} \lambda_{t'} \right) (1-\lambda_t^2)}{1 - \prod_{t'=1}^t \lambda_{t'}^2} \nonumber\\ 
   b^{(t-1)} & =  \frac{\left(1- \prod_{t'=1}^{t-1} \lambda_{t'}^2 \right) \lambda_t}{1 - \prod_{t'=1}^t \lambda_{t'}^2} \nonumber \\
   \sigma^{2}_{t-1|0,t} &= \frac{\left(1- \prod_{t'=1}^{t-1} \lambda_{t'}^2 \right) (1-\lambda^2_t)}{1 - \prod_{t'=1}^t \lambda_{t'}^2}. \label{eq:cond-mean}
\end{align}
}

\subsection{Parameterising the Model}

We have said that a neural network will be used to parameterise $\mu_{\theta}(x^{(t)},t-1)$ and $\sigma_{\theta}^2(x^{(t)},t-1)$. One option would be to use a standard architecture and to parameterise the mean and variance directly without modification. However, we can do better than this by building in an inductive bias that will help the network operate across the different fidelity levels. There are many ways to do this but here we will highlight two options for both the mean and the variance.

\textbf{Mean Option 1: $\bm{x^{(0)}}$-parameterisation}. First we take inspiration from $q(x^{(t-1)}|x^{(0)},\;x^{(t)})$. Specifically, consider the conditional mean (equation \ref{eq:cond-mean}) which tells us how to compute the best guess for $x^{(t-1)}$ if we know $x^{(t)}$ and $x^{(0)}$,
\begin{align}
\mu_{t-1|0,t} &= a^{(t-1)}  x^{(0)} + b^{(t-1)} x^{(t)}. \nonumber\end{align}
We could therefore ask the network to predict the clean data at each step, $\hat{x}^{(0)}_{\theta}(x^{(t)},t-1) \approx x^{(0)}$, and use this to compute the mean
\begin{align}
\mu_{\theta}(x^{(t)},t-1) &= a^{(t-1)}  \hat{x}^{(0)}_{\theta}(x^{(t)},t-1) + b^{(t-1)} x^{(t)} \nonumber.
\nonumber\end{align}
In this way the network is always asked to predict the same thing (the original data) regardless of the augmentation level $t$ we are operating at which reduces how much it has to be adapted from one level to the next.  Notice that this construction also builds in a sensible linear residual connection to $x^{(t)}$ and a rescaling factor. The approach of estimating the clean data from noisy versions relates to the denoising auto-encoder \citep{vincent+al:2008}.\sidenote{\footnotesize 
\textbf{Relationship to the denoising auto-encoder}. The denoising auto-encoder performs representation learning by training a network $h(\cdot)$ to denoise corrupted data $x^{(0)}+\epsilon$ via minimising $\mathbb{E}_{q(x^{(0)})q(\epsilon)} \left[ \left(x^{(0)}-h(x^{(0)}+\epsilon)\right)^2 \right]$. We can compare this to our case. As we have based the parameterisation of $\mu_{\theta}(x^{(t)},t-1)$ on the form of the conditional mean $\mu_{t-1|0,t}$, the squared difference in the objective function now has a simple form
\begin{align}
&\mathbb{E}_{q(x^{(0)},\;x^{(t)})} \Biggl[(\mu_{t-1|0,t} - \mu_{\theta}(x^{(t)},t-1))^2 \Biggl] = \nonumber \\
    &  \left ( a^{(t-1)} \right)^2 \mathbb{E}_{q(x^{(0)},\;x^{(t)})} \Biggl[(x^{(0)} - \hat{x}^{(0)}_{\theta}(x^{(t)},t-1))^2\Biggl].\nonumber\end{align}
    So our network also tries to denoise each $x^{(t)}$ into $x^{(0)}$ as in the denoising auto-encoder (for more information see the Variational Diffusion Model objective in the next section).
}
Notice that the denoising process now depends on the parameters of the noising process through $a^{(t-1)}$ and $b^{(t-1)}$.

\textbf{Mean Option 2: $\bm{\epsilon}$-parametersiation}. Alternatively, we can ask the network to estimate the noise which has been added to $x^{(0)}$ to produce $x^{(t)}$. Using equation \ref{eq:cond-samp} we have\sidenote{\footnotesize Here we have defined
\begin{align}
c^{(t)} & = \prod_{t'=1}^t \lambda_{t'} \; \; \text{and} \;\; d^{(t)}   = \sqrt{1- \prod_{t'=1}^t \lambda_{t'}^2}.\nonumber\end{align} }
\begin{align}
    x^{(t)} & = c^{(t)}  x^{(0)} + d^{(t)} \epsilon^{(t)} \;\; \text{where} \;\; \epsilon^{(t)} \sim \mathcal{N}(0,1).\nonumber
\end{align}
We can substitute this into the expression for the conditional mean
\begin{align}    
\mu_{t-1|0,t} &=  \frac{a^{(t-1)}}{c^{(t)}} (x^{(t)} - d^{(t)} \epsilon^{(t)} )+ b^{(t-1)} x^{(t)}. \nonumber
\end{align}
So we can ask the network to estimate the noise in $x^{(t)}$ i.e.~$\hat{\epsilon}^{(t)}_{\theta}(x^{(t)},t-1) \approx \epsilon^{(t)}$, and parameterise the mean via
\begin{align}    
\mu_{\theta}(x^{(t)},t-1) &=  \frac{a^{(t-1)}}{c^{(t)}} (x^{(t)} - d^{(t)} \hat{\epsilon}^{(t)}_{\theta}(x^{(t)},t-1) ) +b^{(t-1)} x^{(t)} . \nonumber
\end{align}
Again this has a linear (residual) connection to $x^{(t)}$ and a rescaling factor. This approach relates to denoising score-matching \citep{song+ermon:2019}.\sidenote{\footnotesize \label{sidenote:epsilon-param} 
\textbf{Relationship to denoising score matching}. Denoising score matching also considers an augmented dataset $q(x,x^{(0)}) = q(x^{(0)} ) q(x | x^{(0)},\phi)$ where $q(x | x^{(0)},\phi)$ is a noising process with parameters $\phi$ e.g.~Gaussian additive noise. The goal is to build a model for the noised data $x$. The noise is necessary to enable the estimation scheme, but if it is small then $q(x | \phi) \approx q(x^{(0)})$ and the model will be a reasonable approximation to data distribution even though it models the noised data $q(x|\phi)$. 
More specifically, denoising score matching trains a neural network $s_{\theta}(x,\phi)$ to estimate the derivative of the log density of the noised data $s_{\theta}(x,\phi) \approx \frac{\mathrm{d}}{\mathrm{d} x} \log q(x | \phi)$ which is known as the score. The score can then be used to sample from the model using Langevin sampling \citep{mackay:2003}. Estimation of the score is performed by optimising the following cost
\begin{align}
\mathbb{E}_{q(x^{(0)},x|\phi)} \left[ \left(s_{\theta}(x,\phi) - \frac{\mathrm{d}}{\mathrm{d} x} \log q(x | x^{(0)},\phi) \right)^2\right] \nonumber
\end{align}
which has a minimum when
$s^{*}_{\theta}(x,\phi) = \frac{\mathrm{d}}{\mathrm{d} x} \log q(x | \phi)$.
For an AR(1) Gaussian noising process where $q(x | x^{(0)},\phi) = \mathcal{N}(x ; \lambda x^{(0)},\sigma^2)$ then
$\frac{\mathrm{d}}{\mathrm{d} x} \log q(x | x^{(0)},\phi) = - (x-\lambda x^{(0)}) / \sigma^2 = - \epsilon/\sigma $ which means the cost becomes
\begin{align}
 \mathbb{E}_{q(x^{(0)})q(\epsilon)} \left[ \frac{1}{\sigma^2} \left( \left( - \sigma s_{\theta}(x,\sigma^2)\right) - \epsilon  \right)^2\right] \nonumber
\end{align}
Here $\hat{\epsilon}_{\theta}(x,\sigma^2) = - \sigma s_{\theta}(x,\sigma^2) \approx \epsilon$.
Let's compare this cost to the $\epsilon$-parameterisation of the DDPM. Again, the similarity in the parameterisation of $\mu_{\theta}(x^{(t)},t-1)$ and $\mu_{t-1|0,t}$, means that the squared difference in the objective function  has a simple form
\begin{align}
    &\mathbb{E}_{q(x^{(0)},\;x^{(t)})} \Biggl[(\mu_{t-1|0,t} - \mu_{\theta}(x^{(t)},t-1))^2 \Biggl] = \nonumber \\
    & \quad \kappa^{(t)} \mathbb{E}_{q(x^{(0)},\;x^{(t)})} \Biggl[(\epsilon^{(t)} - \hat{\epsilon}^{(t)}_{\theta}(x^{(t)},t-1))^2\Biggl].\nonumber\end{align}
where $\kappa^{(t)} = \left ( a^{(t-1)} d^{(t)}/c^{(t)} \right)^2$. (For full details about the  Simplified DDPM objective in the next section and \citep{song+ermon:2019}.)
By corresponding the two results we see that, at a fixed noise level $t$, the network's estimate  for $\epsilon^{(t)}$ will be related to the score, 
\begin{align}
s_{\theta}(x^{(t)}) & = - \frac{1}{\sqrt{1-\Lambda_t^2}}\hat{\epsilon}^{(t)}_{\theta}(x^{(t)},t-1)\nonumber\\
&\approx  \frac{\mathrm{d}}{\mathrm{d} x^{(t)}} \log q(x^{(t)} |\Lambda_t). \nonumber \end{align} }

\textbf{Variance Option 1}. Learning of the variance $\sigma_{\theta}^2(x^{(t)},t-1)$ has been found to be challenging \citep{nichol+dhariwal:2021}.\sidenote{\footnotesize Although it seems surprising, in light of what follows, that the following option does not work reasonably: 
\begin{align}&\sigma_{\theta}^2(x^{(t)},t-1) = \nonumber\\ & \quad \rho_{\theta}(x^{(t)},t-1) \sigma_t^2 + (1-\rho_{\theta}(x^{(t)},t-1)) \sigma^{2}_{t-1|0,t}\nonumber\end{align} where $\rho_{\theta}(x^{(t)},t-1)$ is a neural network with a sigmoid final layer. } Therefore it is typically set to a constant value which is independent of $x^{(t)}$. One option is to set it equal to the variance of the noising process
\[ \sigma_{\theta}^2(x^{(t)},t-1) = \sigma_t^2 = 1 - \lambda_t^2.
\]
This is optimal when the data themselves are drawn from a unit Gaussian\sidenote{\footnotesize This can be seen intuitively by realising that if $x^{(0)} \sim \mathcal{N}(0,1)$ then the DDPM is initialised with its invariant distribution and so running it from $x^{(T)} \sim \mathcal{N}(0,1)$ to $x^{(0)}$ is equivalent to running it from $x^{(0)}$ to $x^{(T)}$.} or when there are an infinite number of augmentations $T \rightarrow \infty$ \citep{anderson:1983}. However, generally this variance will be overly large.

\textbf{Variance Option 2}. An alternative option again takes inspiration from the conditional distribution $q(x^{(t-1)}|x^{(0)},\;x^{(t)})$. Specifically,  equation \ref{eq:cond-mean}  tells us the variance of $x^{(t-1)}$ if we know $x^{(t)}$ and $x^{(0)}$. We can set the denoising variance to be equal to this, 
\[ \sigma_{\theta}^2(x^{(t)},t-1) = \sigma^{2}_{t-1|0,t}.\]
This is optimal when the data only take a single value (i.e.~it is always known). So generally this will under-estimate the variance.

\subsection{Putting it all together and choosing the schedule}

The DDPM framework we have specified above comprises a family of data augmentations, associated objective functions, and models. On the augmentation side, we have to select a set of augmentation coefficients $\{ \lambda_t \}_{t=1}^T$. On the objective function side we must choose the weights $\{ w_t \}_{t=1}^T$ (or  use an unweighted objective $w_t=1$). On the modelling side, we can select how we parameterise the mean $\mu_{\theta}(x^{(t)},t-1)$, whether we let the variance $\sigma_{\theta}^2(x^{(t)},t-1)$ depend on $x^{(t)}$, and what parameterisation we use for it. Below we will cover two popular choices for the objective function before discussing choices for the augmentation coefficients.

\textbf{Simplified DDPM objective}. First we consider the simplified training objective in \cite{ho+al:2020}. This is recovered from the framework described above when we make the following choices: we use the $\epsilon$-parameterisation of $\mu_{\theta}(x^{(t)},t-1)$, set   $\sigma_{\theta}^2(x^{(t)},t-1) = \sigma_t^2$ and then use $w_{t-1} = \left( \frac{\sigma_t \; c^{(t)}}{a^{(t-1)} d^{(t)}} \right)^2$ so that the weights cancel with the terms multiplying the noise variables (see note \ref{sidenote:epsilon-param}). We then drop constant terms in the objective that do not depend on the parameters $\theta$ to yield,
\begin{align}
\mathcal{L}(\theta ) = - \frac{T}{2} \mathbb{E} \left[     \left({\color{blue} \epsilon^{(t)}} - \hat{\epsilon}^{(t)}_{\theta}(\underbrace{\Lambda_t x^{(0)} + \sqrt{1-\Lambda^2_t} {\color{blue}\epsilon^{(t)}}}_{x^{(t)}},t-1) \right)^2\; \right] \nonumber 
\end{align}
where the expectation is over $t\sim\text{Uniform(1,T)}$ and $(x^{(0)},\;x^{(t)})\sim q(x^{(0)},\;x^{(t)})$, and we have used the shorthand $\Lambda_t = \prod_{t'=1}^t \lambda_{t'}$. 

We have coloured the noise variables to make clear that training involves adding noise to clean data and inputting this corrupted data into a neural network which then tries to estimate the noise that you have added. As mentioned in note \ref{sidenote:epsilon-param}, this training procedure is closely related to denoising score matching \citep{song+ermon:2019}. 

\textbf{Variational Diffusion Model objective}. Second we consider one of the discrete time objectives in \cite{kingma+al:2021}.\sidenote{\footnotesize Although this objective plays a central role in \cite{kingma+al:2021}, in practice they use the $\epsilon$-parameterisation for the experiments.} This is recovered when we make the following choices: we use the $x^{(0)}$-parameterisation of $\mu_{\theta}(x^{(t)},t-1)$, set   $\sigma_{\theta}^2(x^{(t)},t-1) = \sigma_{t-1|0,t}^2$ and use the unweighted objective $w_{t-1} = 1$. Again we drop terms that do not depend on the parameters finding
\begin{align}
\mathcal{L}(\theta ) = - \frac{T}{2} \mathbb{E} \left[   \left(\text{SNR}(t) - \text{SNR}(t-1)\right)  \left(x^{(0)} - \hat{x}^{(0)}_{\theta}(x^{(t)},t-1) \right)^2 \right] \label{eq:vDDPM}
\end{align}
where the expectation is over $t\sim\text{Uniform(1,T)}$ and $(x^{(0)},\;x^{(t)})\sim q(x^{(0)},\;x^{(t)})$. Notice, as mentioned earlier, that this setup relates to the denoising auto-encoder \citep{vincent+al:2008}.

Here we have defined a signal to noise ratio (SNR\sidenote{\footnotesize In more detail we can take the relationship between $x^{(t)}$ and the clean data (equation \ref{eq:cond-samp}) and use this to decompose its the variance, 
 \begin{align}
     x^{(t)} & = x^{(0)}\prod_{t'=1}^{t} \lambda_{t'} + \epsilon_t \left( 1 - \prod_{t'=1}^t \lambda_{t'}^2\right)^{1/2} \nonumber \\
     \text{var}(x^{(t)} ) & = \text{var}\left(x^{(0)}\prod_{t'=1}^{t} \lambda_{t'} \right) + \nonumber \\ & \quad \quad \quad \quad \text{var}\left(\epsilon_t \left (1 - \prod_{t'=1}^t \lambda_{t'}^2 \right)^{1/2}\right) \nonumber \\
     & = \underbrace{\prod_{t'=1}^{t} \lambda_{t'}^2}_{\text{signal variance}} + \underbrace{1-\prod_{t'=1}^t \lambda_{t'}^2}_{\text{noise variance}}.\nonumber
 \end{align}
These are the two terms that are used to define the SNR.}) of level $t$ as the variance in $x^{(t)}$ coming from the raw data $x^{(0)}$ divided by the variance of the noise contribution to $x^{(t)},$
\sidenote{\footnotesize The SNR arises in the objective since 
  \begin{align} \frac{\left(a^{(t-1)} \right)^2}{ \sigma_{\theta}^2(x^{(t)},t-1)} &= \frac{\left(a^{(t-1)} \right)^2}{\sigma^{2}_{t-1|0,t}} \nonumber \\ &= \text{SNR}(t) - \text{SNR}(t-1).\nonumber \end{align}.
  }
\[ \text{SNR}(t) = \frac{ \prod_{t'=1}^{t} \lambda_{t'}^2 }{1 - \prod_{t'=1}^t \lambda_{t'}^2}.\]
Weighting by the difference in the SNRs is natural -- steps where there is a large change in the SNR get more weight.

\textbf{Choosing the schedule}. Finally, we have to select a set of augmentation coefficients $\{ \lambda_t \}_{t=1}^T$. Here we are into a somewhat of a dark art with choices including using a linear schedule for the augmentation noise $1-\lambda^2_t = \beta_1 + (t-1) \beta_2 $ \citep{ho+al:2020},  quarter-cosine schedules for $\Lambda_t$ \citep{nichol+dhariwal:2021}, and linear spacing according to log-SNR \citep{kingma+al:2021}.

One piece of theory we have available to us to guide this choice comes from taking the limit $T \rightarrow \infty $ of equation \ref{eq:vDDPM}. In this limit, the loss does not depend on the schedule --- only the signal to noise ratio at the end-points $t=1$ and at $T \rightarrow \infty$ matter \citep{kingma+al:2021}.\sidenote{\footnotesize See appendix \ref{sec:SNR-loss} for full details. Roughly, the difference in SNRs turns into a derivative w.r.t.~time and the discrete uniform distribution turns into a uniform density. 
Since the SNR is monotonically decreasing by construction (as we add more and more noise as $t$ increases) we can safely reparameterise from time to SNR-level via $u = \text{SNR}(t)$ and use the chain rule to give, 
 \begin{align}
\mathcal{L}(\theta ) & \rightarrow - \frac{1}{2} \int_{u = \text{SNR-min}}^{\text{SNR-max}} \mathbb{E}_x \left[   \left(x^{(0)} - \hat{x}^{(0)}_{\theta}(x^{(u)},u) \right)^2 \right] \mathrm{d}u \nonumber
 \end{align}
 Here $\text{SNR-max}$ is the SNR at level $t=1$ and it is set by $\lambda_1$ and $\text{SNR-min}$ is the SNR at level $T\rightarrow\infty$ (and would typically be 0). We have used overloaded notation so that $\hat{x}^{(0)}_{\theta}(x^{(u)},u) = \hat{x}^{(0)}_{\theta}(x^{(t(u))},t(u)-1)$.
 } So, in this limit, all that matters is the SNR at the start and end of the augmentation and the objective is invariant to the particular path between them. However in practice when approximating the loss via sampling, different schedules can lead to estimators with different variances and strategies to learn schedules which minimise this variance have been developed \citep{kingma+al:2021}.

% \begin{marginfigure}%
%    \includegraphics[width=0.8\linewidth]{figures/input.png}
%    \caption{The input to a transformer is $N$ vectors $\bm{x}_n^{(0)}$ which are each $D$ dimensional. These can be collected together into an array $X^{(0)}$.}
%    \label{fig:input}
%  \end{marginfigure}

\section{Conclusion}

So, to conclude, the DDPM can be broken down into the following steps with the following rationale and design choices:
\begin{enumerate}
\item \textbf{augmentation scheme}: turns generative modelling into a sequence of simple regression problems, choices include Gaussian auto-regression with variance preserving or variance exploding formulations, blurring etc.
\item \textbf{step-wise objective}: used to train the non-linear regression models at each step, typically the choice is to use the maximum-likelihood fitting objective.
\item \textbf{parameter tying and weighted objective}: to reduce the number of parameters of the model and to accelerate training, this requires choosing an architecture which can be shared across augmentation levels. 
\item \textbf{selecting the denoising model}: specification of the regression model architecture, this involves selecting between Gaussian versus non-Gaussian models, with Gaussian models allowing analytic averaging over augmentation data. Other choices for the Gaussian model include whether to use fixed variances, input dependent variances, or correlated noise models.
\item \textbf{parameterisation}: selecting how to use the neural network to parameterise the probabilistic model to make it simple for the network to operate across augmentation levels, choices include the $\epsilon$ and $x^{(0)}$ parameterisations, and also a choice for the variances of the regression models.
\item \textbf{combining the above choices and setting a schedule}: finally we have to put everything together and select a set of augmentation parameters which lead to low-variance estimates of the training objective.
\end{enumerate}

\section{Appendices}
 \subsection{Results for Gaussian AR(1) processes}

This section lays out the rather simple, but laborious algebra that underlies the results that the DDPM uses for the Gaussian AR(1) processes. We start by reminding the reader of the noising process here. 

The process is a joint distribution over the augmented data set
\[
q(x^{(0:T)}) = q(x^{(0)}) \prod_{t=1}^T  q(x^{(t)} | x^{(t-1)})
\]
The highest fidelity level is initialised from the data distribution $x^{(0)} \sim q(x^{(0)})$ where we have assumed that the data have zero mean $\mathbb{E}_{q(x^{(0)})}[x^{(0)}] = 0$ and unit variance $\mathbb{E}_{q(x^{(0)})}[(x^{(0)})^2] = 1$. Practically, as we are assuming a large data set, we can use the empirical mean and standard deviation to normalise the data.

The noising process can then be defined in terms of conditional probability distributions
\[ q(x^{(t)} | x^{(t-1)}) = \mathcal{N}(x^{(t)} ; \lambda_{t} x^{(t-1)},\sigma_t^2), \]
or in terms of samples
\[ x^{(t)} = \lambda_t x^{(t-1)} + \sigma_t \epsilon_t \;\; \text{where} \;\epsilon_t \sim \mathcal{N}(0,1).
 \]
We will use both forms in the derivations that follow.

 \subsubsection{Variance preserving Gaussian AR(1) processes} \label{sec:variance-preserve}
We will now show that using $\sigma_t^2 = 1 - \lambda_t^2$ will ensure that every variable $x^{(t)}$ will be zero mean  and unit variance. This is called the variance preserving AR(1) process.

We will use a recursion to show this --- remember that we already know the data have zero mean  and unit variance so we start by considering $x^{(1)}$. It's relatively simple to show the mean of this variable is zero as it is formed from a linear combination of zero-mean variables 
\[
\mathbb{E}_{q(x^{(1)})}[x^{(1)}] =  \mathbb{E}_{q(x^{(0)},\epsilon_1)} [\lambda_1 x^{(0)} + \sigma_1 \epsilon_1 ]=  \lambda_1 \mathbb{E}_{q(x^{(0)}}[x^{(0)}] + \sigma_1 \mathbb{E}_{q(x^{(0)},\epsilon_1)}[\epsilon_1] = 0
\]
Now consider the variance of $x^{(1)}$
\begin{align}
\mathbb{E}_{q(x^{(1)})}[(x^{(1)})^2] &=  \mathbb{E}_{q(x^{(0)},\epsilon_1)} [(\lambda_1 x^{(0)} + \sigma_1 \epsilon_1 )^2] \nonumber \\
&=  \lambda^2_1 \mathbb{E}_{q(x^{(0)}}[(x^{(0)})^2] + 2 \sigma_1 \lambda_1\mathbb{E}_{q(x^{(0)},\epsilon_1)}[x^{(0)} \epsilon_1] +  \lambda^2_1\mathbb{E}_{q(\epsilon_1)}[\epsilon_1^2] \nonumber \\
& = \lambda^2_1 + 2 \sigma_1 \lambda_1\mathbb{E}_{q(x^{(0)})}[x^{(0)}] \mathbb{E}_{q(\epsilon_1)}[ \epsilon_1] +  \sigma^2_1 \nonumber \\
& = \lambda^2_1 + \sigma^2_1.\nonumber
\end{align}
So if the variance of $x^{(1)}$ is set to unity, then $1 =\lambda^2_1 + \sigma^2_1$ which  means  $\sigma^2_1 = 1 - \lambda^2_1$.

Now that $x^{(1)}$ has zero mean and unit variance we can apply precisely the same argument outlined above to $x^{(2)}$ finding  $\sigma^2_2 = 1 - \lambda^2_2$ , and so on. So, $\sigma_t^2 = 1 - \lambda_t^2$ leads to the variance preserving process.

\subsubsection{Conditioning the Gaussian AR(1) Process on an initial value}\label{sec:condition-init}

We will now show that, for a variance preserving process where $\sigma_t^2 = 1 - \lambda_t^2$, when we condition the process on an initial value $x^{(0)}$ the distribution over $x^{(t)}$ is given by
\[q(x^{(t)}|x^{(0)}) = \mathcal{N}\left(x^{(t)}; \left( \prod_{t'=1}^t \lambda_{t'} \right) x^{(0)}, 1 - \prod_{t'=1}^t \lambda_{t'}^2 \right)\]

To show this, we can unroll the sampling formula for the process
\begin{align}
x^{(t)} = \lambda_t x^{(t-1)} + \sigma_t \epsilon_t = \lambda_t ( \lambda_{t-1} x^{(t-2)} + \sigma_{t-1} \epsilon_{t-1}) + \sigma_t \epsilon_t \nonumber\\
 = \lambda_t ( \lambda_{t-1} (\lambda_{t-2} x^{(t-3)} + \sigma_{t-2} \epsilon_{t-2})) + \sigma_{t-1} \epsilon_{t-1}) + \sigma_t \epsilon_t\nonumber
\end{align}
If we keep unrolling back to $x^{(0)}$ we will get 
\begin{align}
x^{(t)} = &\left (\prod_{t'=1}^t \lambda_{t'} \right) x^{(0)} + \sigma_t \epsilon_t + \lambda_t \sigma_{t-1} \epsilon_{t-1} + \lambda_t \lambda_{t-1} \sigma_{t-2} \epsilon_{t-2} \nonumber \\ 
& \quad \quad \quad \quad \quad \quad \quad + \lambda_t \lambda_{t-1} \lambda_{t-2} \sigma_{t-3} \epsilon_{t-3} + \hdots + \left (\prod_{t'=2}^{t} \lambda_{t'} \right) \sigma_{1} \epsilon_{1}. \nonumber
\end{align}
Remembering that the variance of independent variables, such as $\epsilon_{1:t}$, adds we have
\[q(x^{(t)}|x^{(0)}) = \mathcal{N}\left(x^{(t)}; \left( \prod_{t'=1}^t \lambda_{t'} \right) x^{(0)},  \sigma_{t | 0}^2 \right)\]
where the conditional variance is given by
\[\sigma_{t | 0}^2 = \sigma_t^2 + \lambda^2_t \sigma^2_{t-1}  + \lambda^2_t \lambda^2_{t-1} \sigma^2_{t-2}  + \lambda^2_t \lambda^2_{t-1} \lambda^2_{t-2} \sigma^2_{t-3} + \hdots + \left (\prod_{t'=2}^{t} \lambda^2_{t'} \right) \sigma^2_{1}\]

Now we substitute in the variance preserving condition $\sigma_t^2 = 1 - \lambda_t^2$ and the magic happens: the sum telescopes away to leave only the end points
\begin{align}
\sigma_{t | 0}^2 & = (1 - \Ccancel[blue]{\lambda_t^2}) + \lambda^2_t (\Ccancel[blue]{1} - \Ccancel[red]{\lambda_{t-1}^2})  + \lambda^2_t \lambda^2_{t-1} (\Ccancel[red]{1} - \Ccancel[magenta]{\lambda_{t-2}^2})  + \lambda^2_t \lambda^2_{t-1} \lambda^2_{t-2} (\Ccancel[magenta]{1} - \Ccancel[orange]{\lambda_{t-3}^2)} + \nonumber \\ &\quad  \quad \quad \quad \quad \quad \quad \quad \quad \quad \quad \quad\quad \quad \hdots + \left (\prod_{t'=2}^{t} \lambda^2_{t'} \right) (\Ccancel[black]{1} - \lambda_{1}^2) 
 = 1 - \prod_{t'=1}^t \lambda_{t'}^2\nonumber
\end{align}

\subsubsection{Conditioning on an initial value and the fidelity level below}\label{sec:condition-init+fid}

We will now show that
\[q(x^{(t-1)}|x^{(0)},\;x^{(t)}) = \mathcal{N}(x^{(t-1)}; \mu_{t-1|0,t}, \sigma^2_{t-1|0,t})\]
where
\begin{align}
   \mu_{t-1|0,t} &= a^{(t-1)}  x^{(0)} + b^{(t-1)} x^{(t)} \nonumber\\ 
     & = \frac{\left(\prod_{t'=1}^{t-1} \lambda_{t'} \right) (1-\lambda_t^2)}{1 - \prod_{t'=1}^t \lambda_{t'}^2} x^{(0)} +  \frac{\left(1- \prod_{t'=1}^{t-1} \lambda_{t'}^2 \right) \lambda_t}{1 - \prod_{t'=1}^t \lambda_{t'}^2} x^{(t)} \nonumber \\
   \sigma^{2}_{t-1|0,t} &= \frac{\left(1- \prod_{t'=1}^{t-1} \lambda_{t'}^2 \right) (1-\lambda^2_t)}{1 - \prod_{t'=1}^t \lambda_{t'}^2}. \nonumber
\end{align}

We can prove these results by first applying Bayes' rule to decompose $q(x^{(t-1)}|x^{(0)},\;x^{(t)})$ into a product of two distributions that we know the form of already
\begin{align}
 q(x^{(t-1)} | x^{(0)},x^{(t)}) &\propto q(x^{(t-1)} | x^{(0)}) q(x^{(t)} | x^{(t-1)}) \nonumber \\
 & = \mathcal{N}\left(x^{(t-1)} ; \left ( \prod_{t'=1}^{t-1} \lambda_{t'} \right) x^{(0)},  1- \prod_{t'=1}^{t-1} \lambda^2_{t'} \right) \nonumber \\
 &\quad \quad\quad\quad\quad \times \mathcal{N} \left(x^{(t)} ;   \lambda_{t}  x^{(t-1)},  1- \lambda_t^2 \right).\nonumber
 \end{align}
 
We will now make use of the following identity (which can be shown by comparing the quadratic and linear terms in $x$ inside the exponential on both left and right hand sides)
\begin{align}
\mathcal{N}(x;\mu_1,\sigma_1^2) \times \mathcal{N}(\mu_2;a x,\sigma_2^2) \propto \mathcal{N}\left(x;\mu = \sigma^2 \left(\frac{\mu_1}{\sigma_1^2} + a \frac{\mu_2}{\sigma_2^2} \right),\sigma^2 = \frac{ \sigma_2^2  \sigma^2_1}{ \sigma_1^2  + a^2 \sigma^2_2} \right) \nonumber
\end{align}
and we make the following identifications: $x = x^{(t)}$, $\mu_1 = \left ( \prod_{t'=1}^{t-1} \lambda_{t'} \right) x^{(0)}$, $\sigma^2_1 = 1- \prod_{t'=1}^{t-1} \lambda_{t'}^2$, $\mu_2 = x^{(t)} $, $a = \lambda_t$, and $\sigma^2_2 = 1 - \lambda_t^2$. Substituting in
\begin{align}
\sigma^{2}_{t-1|0,t} & = \sigma^2 = \frac{ \sigma_1^2  \sigma^2_2}{ \sigma_2^2  + a^2 \sigma^2_1} = \frac{ \left ( 1- \prod_{t'=1}^{t-1} \lambda_{t'}^2 \right) (1 - \lambda_t^2)}{(1 - \lambda_t^2) + \lambda_t^2 \left ( 1- \prod_{t'=1}^{t-1} \lambda_{t'}^2 \right)}  \nonumber\\
\mu_{t-1|0,t} & = \mu  = \sigma^2 \left(\frac{\mu_1}{\sigma_1^2} + a \frac{\mu_2}{\sigma_2^2}\right) = \sigma^2\left(\frac{\left ( \prod_{t'=1}^{t-1} \lambda_{t'} \right) x^{(0)}}{1- \prod_{t'=1}^{t-1} \lambda_{t'}^2} + \lambda_t \frac{x^{(t)}}{1 - \lambda_t^2} \right) \nonumber
\end{align}
which simplifies down to the expressions given at the start of this section.

\subsubsection{Relating my notation to standard diffusion notation}\label{sec:notation}

In this section we relate my notation to the notation which is more commonly used in DDPM papers
\begin{align}
\alpha_t &= \lambda_t^2\nonumber\\
\beta_t &= 1-\lambda_t^2\nonumber\\
\bar{\alpha}_t &= \prod_{t'=1}^t \alpha_{t'} = \prod_{t'=1}^t \lambda^2_{t'} = \Lambda_t^2. \nonumber
\end{align}

\subsection{Reparameterising the loss using the SNR}\label{sec:SNR-loss}

Here we take the formula for the variational diffusion loss given in equation $\ref{eq:vDDPM}$, slightly generalised to include weights, and take the limit as $T\rightarrow \infty$.
 \begin{align}
\mathcal{L}(\theta ) & = - \frac{1}{2} \sum_{t=1}^T \mathbb{E}_x \left[  w_{t-1} \left(\text{SNR}(t) - \text{SNR}(t-1)\right)  \left(x^{(0)} - \hat{x}^{(0)}_{\theta}(x^{(t)},t-1) \right)^2 \right] \nonumber \\
& = - \frac{1}{2} \sum_{t=1}^T \mathbb{E}_x \left[   w_{t-1} \frac{\text{SNR}(t) - \text{SNR}(t-1)}{1/T}  \left(x^{(0)} - \hat{x}^{(0)}_{\theta}(x^{(t)},t-1) \right)^2 \right] 1/T \nonumber\\
& \rightarrow - \frac{1}{2} \int_{t=1}^T \mathbb{E}_x \left[ w(t-1)  \frac{\mathrm{d} \text{SNR}(t) }{\mathrm{d}t}  \left(x^{(0)} - \hat{x}^{(0)}_{\theta}(x^{(t)},t-1) \right)^2 \right] \mathrm{d}t \nonumber\\
 \end{align}

Since the SNR is monotonically decreasing by construction (as we add more and more noise as $t$ increases) we can safely reparameterise using $u = \text{SNR}(t)$ and use the chain rule to give

 \begin{align}
\mathcal{L}(\theta ) & \rightarrow - \frac{1}{2} \int_{u = \text{SNR-min}}^{\text{SNR-max}} \mathbb{E}_x \left[ w(u)    \left(x^{(0)} - \hat{x}^{(0)}_{\theta}(x^{(t)},u) \right)^2 \right] \mathrm{d}u \nonumber\\
 \end{align}
 Here $\text{SNR-max}$ is the SNR at level $t=1$ and it is set by $\lambda_1$ and $\text{SNR-min}$ is the SNR at level $T\rightarrow\infty$ which is set to $\Lambda_T = \prod_{t=1}^T \lambda_t$ (and would typically be close to 0). We have used overloaded notation so that e.g.~$w(u) = w(t(u)-1) $. 
 
Crucially, this equation does not depend on the schedule --- only the starting and ending signal to noise ratios.

\subsection{Relationship to the Evidence Lower Bound (ELBO)}\label{sec:ELBO}

The DDPM can be viewed as a latent variable model for observed data $x^{(0)}$ with latents $x^{(1:T)}$. The marginal probability density of a data point is given by $p(x^{(0)} | \theta ) = \int p(x^{(0:T)} | \theta ) \;\mathrm{d} x^{(1:T)}$. As this expression measures how probable each data point is under the model, it would make for a sensible training objective for the model parameters $\theta$ e.g.~using maximum-likelihood training which would average the log-density over the training data\sidenote{The average log-likelihood of the parameters with respect to the observed data $\mathbb{E}_{q(x^{(0)})} \left [ \log p(x^{(0)} | \theta ) \right]$ would arguably be a preferable objective to the per-step average log-likelihoods used above $\mathbb{E}_{q(x^{(1:T)})} \left [ \log p(x^{(t-1)} | x^{(t-1)},\theta ) \right]$ as it measures the quality of the entire model on the observed data, rather than measuring the quality of part of the model on augmented data.} 
\[ \theta^* = \argmax_{\theta} \mathbb{E}_{q(x^{(0)})} \left [ \log p(x^{(0)} | \theta ) \right]. \]
Unfortunately $p(x^{(0)} | \theta )$ is intractable and so an approximation is required. The Evidence Lower BOund (ELBO) is one such approximation. It is a lower bound on the likelihood of the parameters for a latent variable model,
\begin{align}
\mathbb{E}_{q(x^{(0)})} \left [ \log p(x^{(0)} | \theta ) \right] \ge \mathbb{E}_{q(x^{(0)})} \left [  \text{ELBO}(x^{(0)},\theta,q(x^{(1:T)})) \right]. \nonumber
\end{align}
The ELBO is formed by subtracting a KL divergence term from the intractable log-likelihood. The KL between two densities $q(x)$ and $p(x)$ is defined as \[\text{KL}( q(x) || p(x) ) = \int q(x) \log \frac{q(x)}{p(x)} \;\mathrm{d} x \ge 0. \]
The KL is non-negative and takes the value zero when $q(x) = p(x)$.
In our case, $p(x)$ will be selected so that when the KL is subtracted from the log-likelihood, the resulting ELBO is tractable. It turns out that choosing it to be  the posterior distribution over the latent variables given the observed variables $p(x) = p(x^{(1:T)}|x^{(0)})$ leads to a simple form. We will talk about the form of $q(x^{(1:T)})$ in a moment (for one thing, we will let it depend on the data $x^{(0)}$), but at this point it can be considered arbitrary. With this setup, the ELBO can be written as follows
\begin{align}
 \text{ELBO}(x^{(0)},\theta,q(x^{(1:T)})) & = \log p(x^{(0)} | \theta ) - \text{KL}( q(x^{(1:T)}) || p(x^{(1:T)}|x^{(0)},\theta) ) \nonumber \\
 & = \int q(x^{(1:T)}) \log \frac{p(x^{(0:T)} | \theta )}{q(x^{(1:T)})} \;\mathrm{d} x^{(1:T)} \nonumber\\
 & = \mathbb{E}_{q(x^{(1:T)})} \left[ \log \frac{p(x^{(0:T)} |\theta)}{q(x^{(1:T)})} \right]. \nonumber
\end{align}
Notice that the choices made above for the KL and the distributions within it result in us only requiring the joint distribution for the model $p(x^{(0:T)} |\theta) = p(x^{(T)}) \prod_{t=1}^T p(x^{(t-1)} | x^{(t-1)},\theta)$ which is how we specified the model in the first place -- there is no longer a requirement for us to compute intractable marginalised quantities. 

Typically, variational inference then involves computing an estimate of the average ELBO by simple Monte Carlo sampling,  
\begin{align}
\mathbb{E}_{q(x^{(0:T)})} \left[ \log \frac{p(x^{(0:T) } | \theta)}{q(x^{(1:T)})} \right] \approx \frac{1}{N} \sum_{n=1}^N \log \frac{p(x_n^{(0:T)} |\theta)}{q(x_n^{(1:T)})} \;\; \text{where} \;\; x_n^{(0:T)} \sim q(x^{(0:T)}). \nonumber
\end{align}
This estimator can then be optimised with respect to the parameters $\theta$. Normally the approximate posterior $q(x^{(1:T)}) = q(x^{(1:T)} | \phi)$ would be parameterised and optimised along side the generative model so that the bound remains as tight as possible at each point, thereby minimising the discrepancy between the ELBO and the log-likelihood \citep{kingma+welling:2022}. After the optimisation the approximate posterior depends on the data $q(x^{(1:T)} | \phi(x^{(0)}))$. This is not the approach used in the DDPM. Instead, 
to recover the DDPM, we use an approximate posterior distribution which depends on the data, but whose parameters are fixed. Specifically, this approximate posterior is equal to the noising process
\[ q(x^{(1:T)} | x^{(0)},\phi) = \prod_{t=1}^T q(x^{(t)} | x^{(t-1)},\lambda_t) = \prod_{t=1}^T \mathcal{N}(x^{(t)}; \lambda_t x^{(t-1)},1-\lambda_t^2).\]
So this approximate posterior distribution depends on the data as the noising process is initialised at $x^{(0)}$, but its parameters $\phi =\lambda_{1:T}$ are not learned.

We can now show that the ELBO objective is equivalent to the unweighted DDPM training objective. First note that 
\begin{align}
&\mathbb{E}_{q(x^{(0)})} \left [  \text{ELBO}(x^{(0)},\theta,q(x^{(1:T)}|x^{(0)},\phi)) \right]
 = \nonumber \\
 & \quad \quad \quad \quad \quad \quad \quad \quad \quad \quad\underbrace{\mathbb{E}_{q(x^{(0:T)}|\phi)} \left [ \sum_{t=1}^T \log p(x^{(t-1)} | x^{(t)}, \theta) \right]}_{\text{unweighted DDPM loss}} + c(\phi) \nonumber
\end{align}
where the constant $c(\phi)$ is the average entropy of the augmented variables and does not depend on the model parameters $\theta$,
\begin{align}
c(\phi) & = - \mathbb{E}_{q(x^{(0:T)}|\phi)} \left [ \log \left ( q(x^{(1:T)}|x^{(0)},\phi) \right) \right].\nonumber
\end{align}
As such optimisation of the ELBO w.r.t.~$\theta$  is equivalent to optimising the unweighted DDPM objective.

\subsubsection{Why I don't like the ELBO perspective}

I don't particularly like the ELBO perspective for the following reasons

\begin{enumerate}
\item  The ELBO perspective immediately suggests generalisations to the DDPM which do not work well including learning the approximate posterior (i.e.~learning the noising process) and using a non-linear noising process parameterised by a neural network. Both of these extensions are natural from the ELBO and VAE perspectives as they will lead to tighter bounds on the log-likelihood. The fact that they do not work well is a challenge to the utility of the ELBO perspective.\sidenote{\footnotesize Part of the reason that variational methods do not work well in general though is that often the optimisation takes us to parameter settings $\theta$ where the bound is tight ($q$ is accurate) rather than where the underlying likelihood is high ($\theta$ is close to the underlying maximum likelihood parameter). So, counterintuitively, there are cases where simpler approximate posteriors perform better e.g.~an approximation which is equally tight across the parameter space is optimal from a parameter-learning perspective (see figure \ref{fig:bounds} and  \cite{turner+sahani:2011a} for more information). This could be used to justify  a fixed and simple $q$, but this is a nuanced justification.}  
\begin{marginfigure}%
\includegraphics[width=1\linewidth]{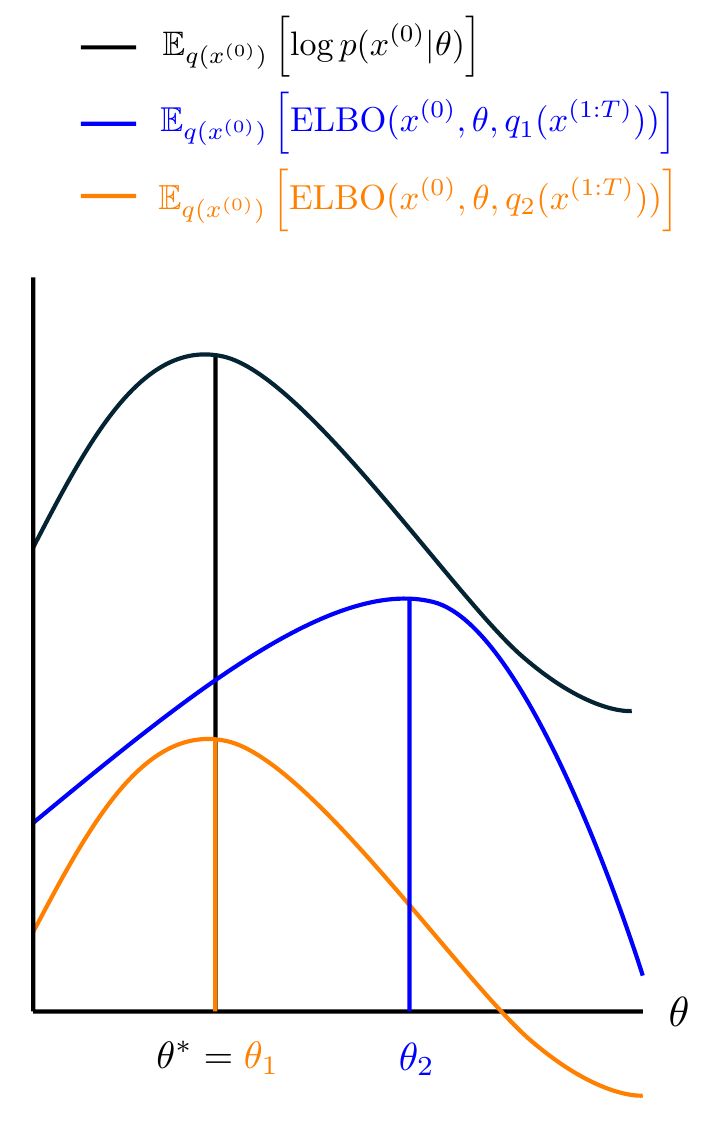}
   \caption{Here's a toy example where the log-likelihood is shown in black with an optimum $\theta^*$. Below it are two different variational lower bounds (ELBOs) formed from using different approximate posteriors $\color{blue}q_1$ and $\color{orange}q_2$. These two bounds have different optima $\color{blue}\theta_1$ and $\color{orange}\theta_2$. The first bound is tighter to the log-likelihood, but its optimum is biased towards large parameter values as the bound is tighter there. The second bound is looser, but as it is equally loose everywhere, its optimum coincides with the log-likelihood. For this reason, simpler families of approximate posterior distribution can sometimes outperform more complex ones as their tightness can be less parameter dependent. 
   \label{fig:bounds}}
 \end{marginfigure}  

\item In many practical situations, authors use the weighted version of the DDPM loss $w_t \ne 1$ to improve performance. In which case, the clean correspondence to the ELBO is lost (the objective is no longer a bound) and again the ELBO does not usefully constrain the design space.\sidenote{\footnotesize For the same reason, in variational auto-encoders, the fact that KL reweighting techniques are necessary to obtain strong performance are a reason to doubt the primacy of the variational lower bound perspective.} So, not only does the ELBO suggests generalisations that do not perform well (point 1 above), it is also incompatible with generalisations that do perform well.

\item If you fix the approximate posterior to a simple distribution and are just learning the generative model, then the ELBO perspective does not bring substantial additional technical insights or utility (although see point 5 below for an exception). However, it comes at the cost of needing to know about KL divergences, bounds etc. In this light, I prefer the simpler explanation given in the main body of this note.

\item  In the normal exposition of probabilistic modelling, latent variables are a hierarchy of  high- and low-level variables that underlie the data. For example, for images they might comprise the objects present in a scene, their pose, position and properties, the lighting conditions, camera position and so on. These types of variables necessitate complex learned inference networks $q$. Fixing the inference network and using a simple a linear Gaussian form would make no sense if the goal is to extract such variables. However, here in the DDPM the latent variables have quite a different character --- they are just noise corrupted versions of the data rather than meaningful high-level representations. It is remarkable from the standard latent variable modelling perspective that latent variables which are representationally meaningless can be used to produce a good generative model. This feature of the DDPM means connections to latent variable modelling are conceptually dissonant, although mathematically accurate.

\item In general an ELBO formed with a linear Gaussian approximate posterior would be extremely loose and a poor optimisation target. This will be true for the DDPM for small to moderate numbers of noise levels $T$. Remarkably, in the limit $T \rightarrow \infty$ though, there exist non-linear generative models that are universal approximators which also have optimal Gaussian AR(1) approximate posteriors \citep{anderson:1983}. It seems likely in this limit that the ELBO will be tight to the log-likelihood, although we do not know of a proof at this time.
%\sidenote{TO DO: It would be good to flesh this argument out in more detail. As a practical test it would be enlightening to compare the likelihoods obtained from the equivalent ODE probability flow equations to the ELBO to check these agree. It would also be good to verify that the two variance parameterisations listed above in section 2.5 are equal in the limit $T \rightarrow \infty$ (this is necessary I think for the KL to remain finite when using the second parameterisation used by the variational diffusion model).} 
This behaviour is very surprising, and essential to understand why the variational approximation is sensible, but again this requires substantial technical explanation and the continuous time version of the DDPM.

%\item  I don't understand why \cite{kingma+al:2021} can use the wrong noise in the model as compared to \cite{anderson:1983} and then get a finite ELBO. Are these equal in the limit?
\end{enumerate}

\vspace{3mm}
{\bf Acknowledgements.} We would like to thank Jos\'e Miguel Hern\'andez-Lobato for helpful feedback. Richard E.~Turner is supported by Microsoft, Google, Amazon, ARM, Improbable and EPSRC grant EP/T005386/1 together with Aliaksandra Shysheya.
%\normalsize

\bibliographystyle{plainnat}
\bibliography{bibliography}

\end{document}